\documentclass[cisrama4paper, 10 pt, conference ]{ieeeconf}  % Comment this line out if you need a4paper
\IEEEoverridecommandlockouts                              % This command is only needed if 
\usepackage{graphics} % for pdf, bitmapped graphics files
\usepackage{epsfig} % for postscript graphics files
\usepackage{mathptmx} % assumes new font selection scheme installed
\usepackage{times} % assumes new font selection scheme installed
\usepackage{amsmath} % assumes amsmath package installed
\usepackage{amssymb}  % assumes amsmath package installed
\usepackage[T1]{fontenc}

\title{\LARGE \bf
	Augmented-Reality-Based Visualization of Navigation Data of Mobile Robots on the Microsoft Hololens - Possibilities and Limitations
}

\author{Linh K{\"a}stner$^{1}$ and Jens Lambrecht$^{1}$% <-this % stops a space
\thanks{$^{1}$Linh K{\"a}stner and Jens Lambrecht are with the Chair Industry Grade Networks and Clouds Department, Faculty of Electrical Engineering, and Computer Science,
        Technical University of Berlin, Berlin, Germany
        {\tt\small linhdoan@tu-berlin.de}}%
}

\begin{document}

\maketitle
\thispagestyle{empty}
\pagestyle{empty}

%%%%%%%%%%%%%%%%%%%%%%%%%%%%%%%%%%%%%%%%%%%%%%%%%%%%%%%%%%%%%%%%%%%%%%%%%%%%%%%%
\begin{abstract}
	
The demand for mobile robots has rapidly increased in recent years due to the flexibility and high variety of application fields comparing to static robots. 
To deal with complex tasks such as navigation, they work with high amounts of different sensor data making it difficult to operate with for non-experts. To enhance user understanding and human robot interaction, we propose an approach to visualize the navigation stack within a cutting edge 3D Augmented Reality device - the Microsoft Hololens. Therefore, relevant navigation stack data including laser scan, environment map and path planing data are visualized in 3D within the head mounted device. Based on that prototype, we evaluate the Hololens in terms of computational capabilities and limitations for dealing with huge amount of real-time data. Results show that the Hololens is capable of a proper visualization of huge amounts of sensor data. We demonstrate a proper visualization of navigation stack data in 3D within the Hololens. However, there are limitations when transferring  and displaying different kinds of data simultaneously.

\end{abstract}

%%%%%%%%%%%%%%%%%%%%%%%%%%%%%%%%%%%%%%%%%%%%%%%%%%%%%%%%%%%%%%%%%%%%%%%%%%%%%%%%
\section{INTRODUCTION}
The usage of mobile robots has increased in recent years within all sectors of society including industries due to their flexibility and the variety of use cases they can operate in. The future factories will rely on the help of mobile robots for tasks such as procurement of components, transportation or commissioning \cite{smartfactory} \cite{mobilerob}. Due to their complexity, operation and understanding of mobile robots is still a privilege to experts \cite{mobilerob2}. Nonetheless, understanding and operation, not just for researchers and developers, but also employees with less technical understanding or prior instruction is paramount in a future environment to make interaction and collaboration processes more time- and cost-efficient, easy to operate and safe \cite{smartfactory2}. Therefore, Augmented Reality (AR) had been subject to various scientific publications which proved the great potential and ability to enhance efficiency in human-robot-interaction (HRI), -collaboration and support understanding \cite{hashimoto}, \cite{heydari}, \cite{fang}.  
AR has the potential to aid the user with help of spatial information and the combination with intuitive interaction technology, e.g. gestures \cite{arsystems}. Our previous work focused on AR-based simplification of robot programming with the help of visualizing spatial information and intuitive gesture commands \cite{lambrecht}. Other work used AR for enhanced visualization of robot data or for multi modal teleoperation \cite{ar1} \cite{armultimodal}. By reason of computational power, most work provide AR with handheld devices or external monitors in 2D. However, the advantages of 3D visualization have proven to be more intuitive and support understanding even further compared to 2D display \cite{fuchs} \cite{3d2d1}. \\
Head mounted displays (HMDs) are considered to visualize 3D data directly within the users gaze for spatial display of information which supports understanding. Furthermore, HMDs bear the advantage that the user will have both hands free for other tasks. This could, for instance, help surgeons to display relevant data into the gaze while still doing the surgery.
Regarding mobile robotics, navigation is one important aspect to consider. It is relying to work with numerous kinds of sensor data including laser scan and the environment map. The visualization of 3D sensor data for robot navigation within a HMD will increase user understanding, reduce operation cycles and make monitoring and maintenance more effective.\\
One of the main bottlenecks of state of the art HMDs is their limited computation power. Especially the visualization of data used for mobile robot navigation is a demanding task as real-time data such as laser scan continuously change, making visualization more demanding.
On this account, this papers motivation is to evaluate the possibility of displaying navigation stack data on a state of the art head mounted AR device within the real operation environment of mobile robot. Therefore, we propose a way to display relevant navigation stack data into the Microsoft Hololens. Finally, we evaluate the Hololens in terms of computational capabilities to give insight whether it is capable of appropriately visualizing navigation stack sensor data for mobile robots.\\
The rest of the paper is structured as follows. Sec. II will give an overview of related work. Sec. III will present the conceptional design of our approach while sec. IV will describe the implementation of the prototype with a demonstration in In Sec. V. Sec. VI describes the experimental setup followed by Sec. VII where the results will be evaluated. Finally, Sec. IX will give a conclusion.

\section{RELATED WORK}
Due to the huge potential of AR, integration into various processes and scenarios was evaluated. Hashimoto et al. \cite{hashimoto}  proposed a teleoperation use case for mobile robots using gesture commands on a tablet. The researchers conducted a study where users controlled a robot with gestures to solve tasks. The study concludes an enhancement in understanding and simplified operation when using the proposed AR prototype to control the robot with gesture commands compared to conventional methods. However, participants requested richer data visualization in order to enhance robot understanding even further. Webel et al. \cite{webel} introduced an AR application for training in maintenance and assembly tasks and conducted a case study to observe enhancements of AR based training of manufacture workers compared to no AR training beforehand. Results show that AR has great potential in reducing error rates and performance time due to tactile feedback and rich virtual information display for guidance. As a matter of computational capabilities, the majority of aforementioned  work achieved AR through handheld devices or external monitors in 2D. 
Various work including the work from Fuchs et al. \cite{fuchs} and Velayutham et al. \cite{3d2d1} have shown the advantages of 3D visualization compared to 2D. Fuchs et al. proved that through a study for medical surgery, where participants using 3D visualization of instruction sets were considerably faster and more accurate. Velayutham et.al. concludes that 3D visualization for medical surgery results in faster operation times due to better understanding. 
Head mounted devices (HMDs) emerged in recent years providing the possibility for 3D data display into the real environment. Furthermore, they bear the advantage of leaving the user with both hands free to operate with other tasks while visualization of spatial information is directly displayed in the user gaze. 
As a first stand alone AR-HMD, the Hololens was introduced in 2016 by Microsoft. It contains an own CPU as well as a holographic processing unit (HPU) which comes along with 2 Gb of RAM and 64 Gb of flash memory. Because of its possibility to work remotely without any external entity, the Hololens has been widely considered for integration to enhance HRI. Furthermore, this bears advantages such as freedom in navigation or independent operation.
Previous work including the work from Liu et al.  \cite{holoperform}, Coppens \cite{holo1} and Vassallo et al. \cite{holograms}  have evaluated general technical aspects of the Hololens such as the accuracy of spatial mapping or gaze gesture commands. Vassallo et al. focuses on evaluation of hologram placement accuracy and spatial mapping. They concluded big potential to work in industrial setups. 
Guhl et al. \cite{guhl} proposed a framework design to work with robots and Head Mounted 3D AR devices. The challenge to establish communication between the two systems, Head mounted device and robot, was achieved by using a 'middleman' in-between instance to transfer information between the two entities. 
Krupke et al. \cite{krupke} proposed a multi modal framework using the \textit{ROSSharp} framework to achieve communication between the robot and the AR device without an external entity. This is demonstrated for the UR5 and Hololens to work with gaze gesture. The authors prove the enhancement and advantages of using HMDs to work with robots. In the context of robot sensor data visualization only Thorstensen \cite{thorstensen} dealt with visualization within an AR device. The author was using the HTC Vive to stream 2D camera frames into the HMD and observed positive effects on user understanding. 
Sauer et al. \cite{surgery} proposed an integration of the Hololens for application in surgical interventions. The authors displayed 3D anatomical models directly to real organs and showed high potential for improvement of surgeons actions. 
In the area of data visualization for sensor data of mobile robots, no paper was found which evaluates the capabilities of the Hololens.
Considering the state of the art in head mounted device integration into robotics for sensor data visualization, this papers purpose is to evaluate the possibilities and limitations of the Hololens as a stand alone AR device for sensor data visualization of mobile robots. This is done in the context of visualizing the navigation stack for better user understanding.

\section{CONCEPTION DESIGN}
Like stated before, the main purpose of this paper is to evaluate the Hololens in terms of the capability in visualizing the navigation stack which contains huge and continuously changing sensor data like laser scan data, the environment map and path planing information. This is done in the context of a more intuitive human robot interaction and better robot understanding. Following data are key parameters for the navigation stack: first, laser scan data, second environment data and third navigation data. Laser scan data is evaluated for position and obstacle tracking. This is done in combination with an internal robot map. The robot continuously evaluates its laser scan data with the existing internal map to localize itself. For the global and local path planing, the combination of laser scan and map is used. Referring to the work from Thorstensen \cite{thorstensen}, visualizing laser scan data proved to help user understanding. The internal robot map should be visualized to provide additional understanding of the robot environment for obstacle awareness and to aid in navigation planning.
Monferrer et al. \cite{monferrer} concludes that visualizing navigation path data improves quality of human robot interaction and understanding because future robot movement is being spatially displayed. In our case, navigation data includes the path the robot will take. This aids the user in understanding future robot intentions which is of great importance e.g. in environments with high amount of robots. For a correct visualization, two important things have to be considered: first the correct alignment of the different coordinate systems of both entities, robot and Hololens. Second, the data transmission from ROS to Hololens and vice versa.

\subsection{Hardware Setup}
The hardware setup contains a mobile robot and a head mounted AR device - the Microsoft Hololens. We are working with a Kuka mobile Youbot running ROS Hydro on a Linux version 12.04 distribution. As a development environment for the Hololens, we use a windows notebook with Unity3D 2018 installed on which the application was developed. All preprocessing of sensor data is done directly at ROS site. Both entities, robot and Hololens, are connected to the same network and communicate through this via a WiFi connection. 

\subsection{Coordinate System Alignment}
\begin{figure}[!h]
	\centering
	\includegraphics[width=3.3in]{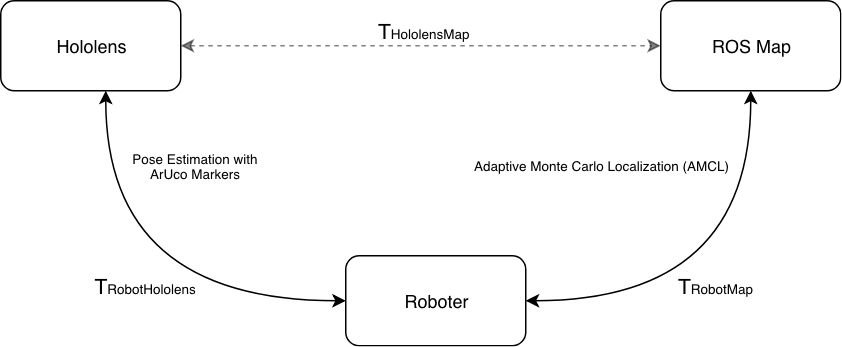}
	\caption{Conception of Coordinate System Alignment}
	\label{conceptcoord}
\end{figure}

The majority of previous work including \cite{krupke} and \cite{guhl} relied on marker detection for pose estimation of the robot. However, for our case, the mobile robot can also move therefore it is necessary to ensure continuous localization of the robot. The main challenge is that both entities are mobile and a single marker tracking approach like used in similar works is not enough. As stated before, to work with the navigation stack, a map is needed. Thus, as a third entity the ROS Map produced by RVIZ has to be considered as well because all sensor data which will be retrieved from ROS topics will be with respect to the map coordinate system. We propose an alignment composed of 2 separate steps: an initial alignment of the robot and the Hololens with a marker detection and after that, continuous alignment with the spatial anchor capability of the Hololens. For this, following formula is to be considered:
\begin{align}
T_{HoloMap} = T_{HoloRobo} \cdot T_{RoboMap}
\end{align} $T_{HM}$ is the transformation of the Hololens position and map. $T_{HR}$ is the pose estimation between the robot and the Hololens and $T_{RM}$ is the transformation of the robot to the origin of the map. The conception is depicted in fig. \ref{conceptcoord}.

\section{IMPLEMENTATION}
\subsection{Communication}
Since ROS and the Hololens are working from different operating systems, an appropriate communication for message exchange is to be considered. Therefore, the Rosbridge protocol for external communication was used which follows a specific JSON notation for message exchange.
\begin{figure}[h!]
	\centering
	\includegraphics[width=3.4in]{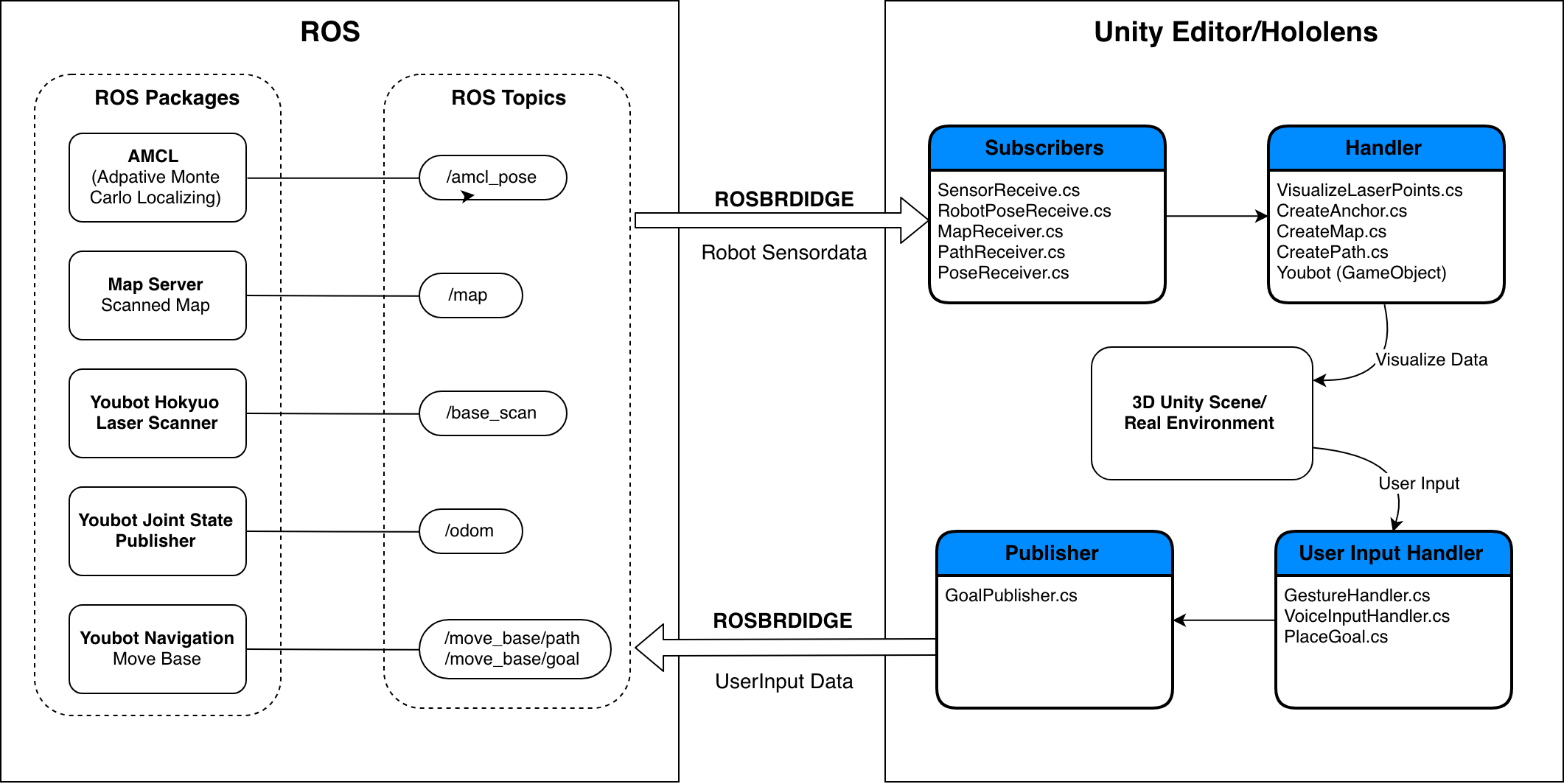}
	\caption{ROS topics and hololens handling scrips}
	\label{tops}
\end{figure}

An open source project of Siemens called \textit{ROSSharp} \cite{bischoff} was used to simplify the process. It provides a framework for seamless communication between the two systems using web sockets. It uses specific publisher and subscriber classes which will publish/subscribe to ROS topics to send and publish messages in the required JSON format of Rosbridge. Handling scripts can be attached to the classes to further process the received data. The topics being used in this work as well as the associated handling scripts are depicted in fig. \ref{tops}.

\subsection{Coordinate Systems alignment}

Like stated before, the challenge is to align the two different coordinate systems of the robot and the Hololens together to ensure correct visualization of all data continuously. Since both entities can move, we propose an alignment composed of two steps: initial pose detection with \textit{ArUco} markers \cite{aruco} and second, using the spatial anchor capability of Hololens to place a virtual anchor for continuous alignment of Hololens and map. 
The Hololens provides the capability of spatial anchors which can memorize the exact position of every point in the environment even after termination of application. This is done with an internal SLAM algorithm based on the multiple environment cameras. The internal SLAM starts by scanning the whole environment with startup of the Hololens. The anchor should act as a common reference point and is placed based on the transformation of Robot and map $T_{RM}$.
For \textit{ArUco} marker tracking to be integrated into Hololens we used the open source implementation of \cite{keymaster}. This gives an appropriate marker tracking for initial pose alignment between robot and Hololens $T_{RH}$. For $T_{RM}$ , we use the \textit{Adaptive Monte Carlo Localization} package of ROS which provides the exact position of the robot with respect to the robot map. On that position the spatial anchor will be placed, providing a reference point for the Hololens. 

\subsection{Sensor Visualization}
After ensuring correct visualization through coordinate system alignment, the sensor data can be visualized within the Hololens. For that, data from ROS has to be converted, interpreted and visualized by the Hololens. The application is developed in Unity. To visualize the incoming data, 3D shapes, the so called \textit{game objects} were used within the Unity. We used 3D spheres to visualize all data. A back end is used for converting raw sensor data from the laser scanner of the Youbot. This is done directly at ROS site. First, the raw laser scan data is visualized by writing an Unity script which interprets the angles and corresponding ranges and generate a game object at all positions to visualize the laserscan data. Therefore the so called meshes were used in unity for efficient processing. 
The environment map of RVIZ provides an occupancy grid which must be converted into a point cloud. Since occupancy grid data only provides 3 different values providing whether each pixel is occupied or not.  A script to interpret and transform the values into a point cloud has been written and executed at ROS site. For the exact position, the size of the whole map is needed which can be extracted from the \textit{yaml} files that comes along when creating RVIZ maps. For our case, the map size was 1060x448 pixels and each pixel represent 0.02 meters.  The preprocessed data is then sent to the Hololens for visualization. 
\begin{figure}[h]
	\centering
	\includegraphics[width=3.3in]{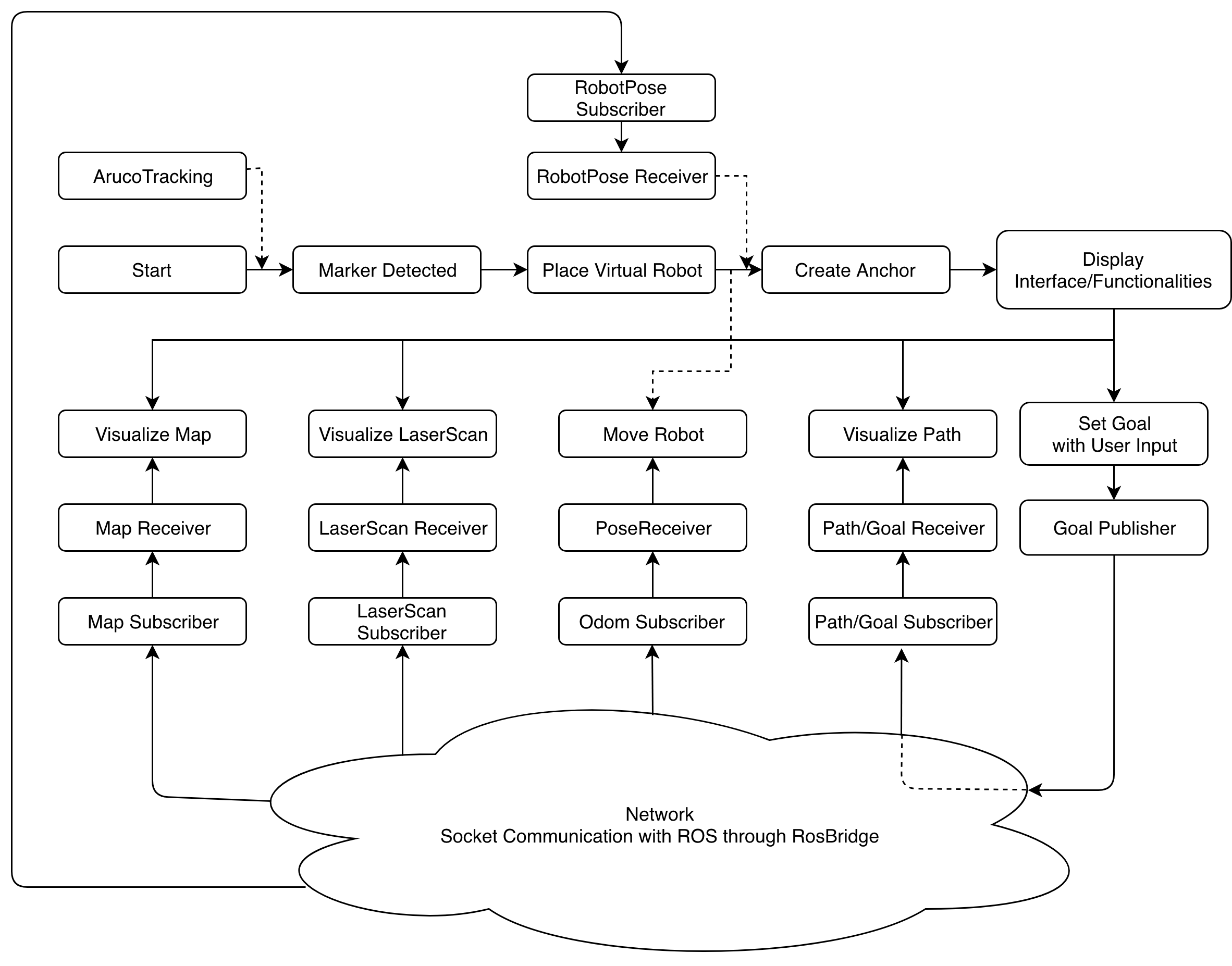}
	\caption{Application Workflow}
	\label{workflow}
\end{figure}
The Navigation data is visualized in form of navigation path of the robot. This information is extracted from the global path planer topic of the Youbot navigation stack. The path is divided into fewer positions and put into an array which is then send to the Hololens for visualization. \\
Fig. \ref{workflow} depicts the workflow of the whole application. First, the initial alignment of Hololens and robot is done via ArUco marker tracking placing an virtual robot at the real one. Afterwards a spatial anchor is placed based on information extracted from the robot pose subscriber which subscribes to Adaptive Monte Carlo localization topic thus aligning the two coordinate systems for further visualization. After these initial steps the data visualization and robot movement control can be executed using the build in interactive user interface containing virtual buttons for each sensor type. For robot navigation the user input is published to the navigation topic. Robot movement is tracked to the odometry topic and based on that the virtual robot moves along with the real one.
\section{PROTOTYPE} 
\begin{figure}[!h]
	\centering
	\includegraphics[width=3.3in]{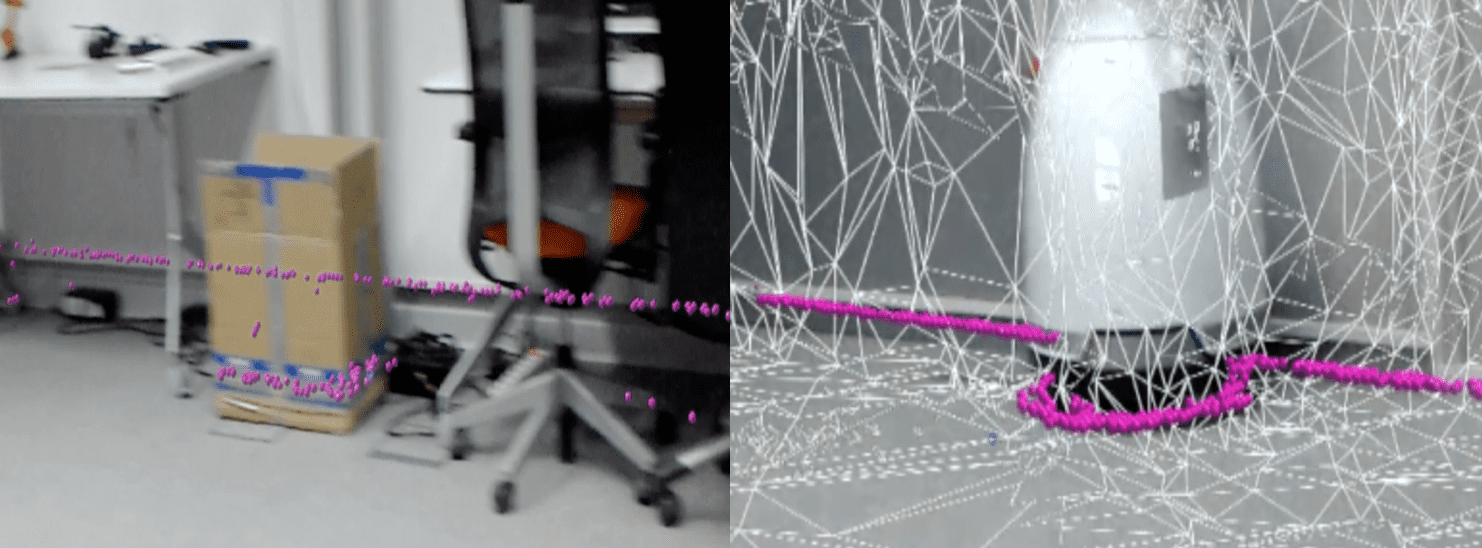}
	\caption{Environment Map Visualization}
	\label{environ}
\end{figure}
\begin{figure}[h!]
	\centering
	\includegraphics[width=3.3in]{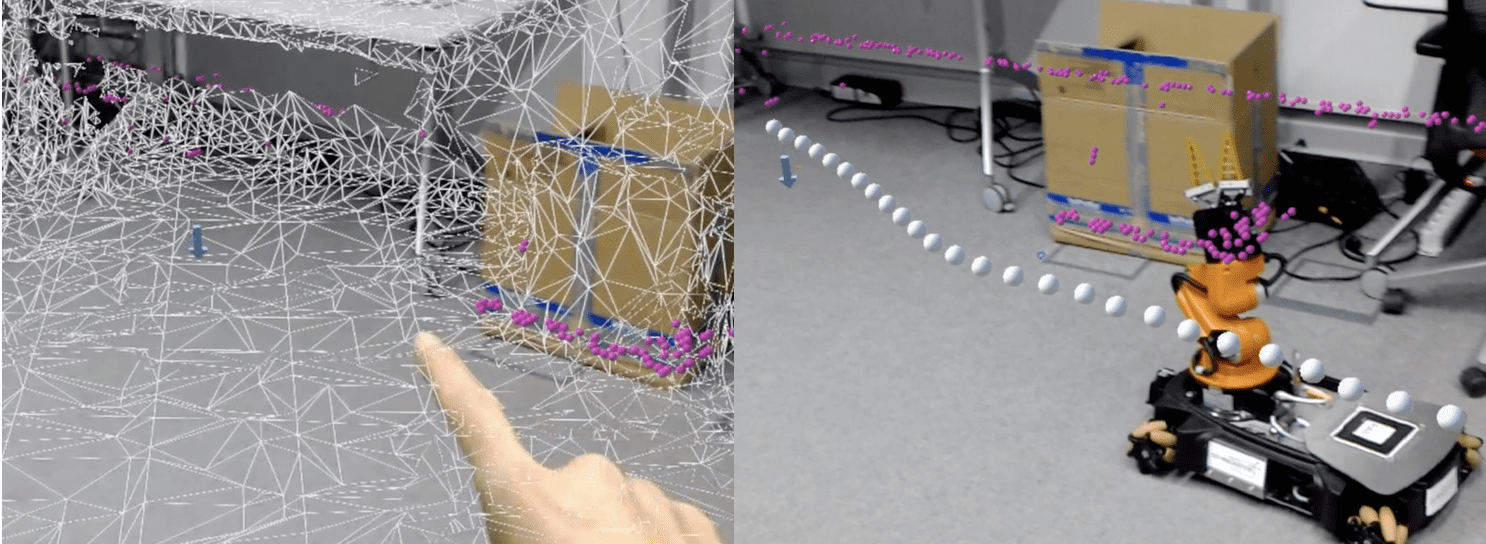}
	\caption{Navigation Goal Definition and Path Visualization}
	\label{goal}
\end{figure}

This section is demonstrating the implemented prototype. Different operation modes of sensor data visualization are shown to validate the functionality of the prototype. Fig. \ref{environ} shows the visualization of the environment map while Fig. \ref{goal} shows the path visualization after placement of a goal position. A goal position is defined by dragging the blue arrow to a location within the room as illustrated in fig. \ref{goal}. For that the spatial mapping capability of the Hololens is turned on to determine possible placement locations within the room. After goal definition, the location is send to ROS triggering robot movement.

\begin{figure}[h!]
	\centering
	\includegraphics[width=3.3in]{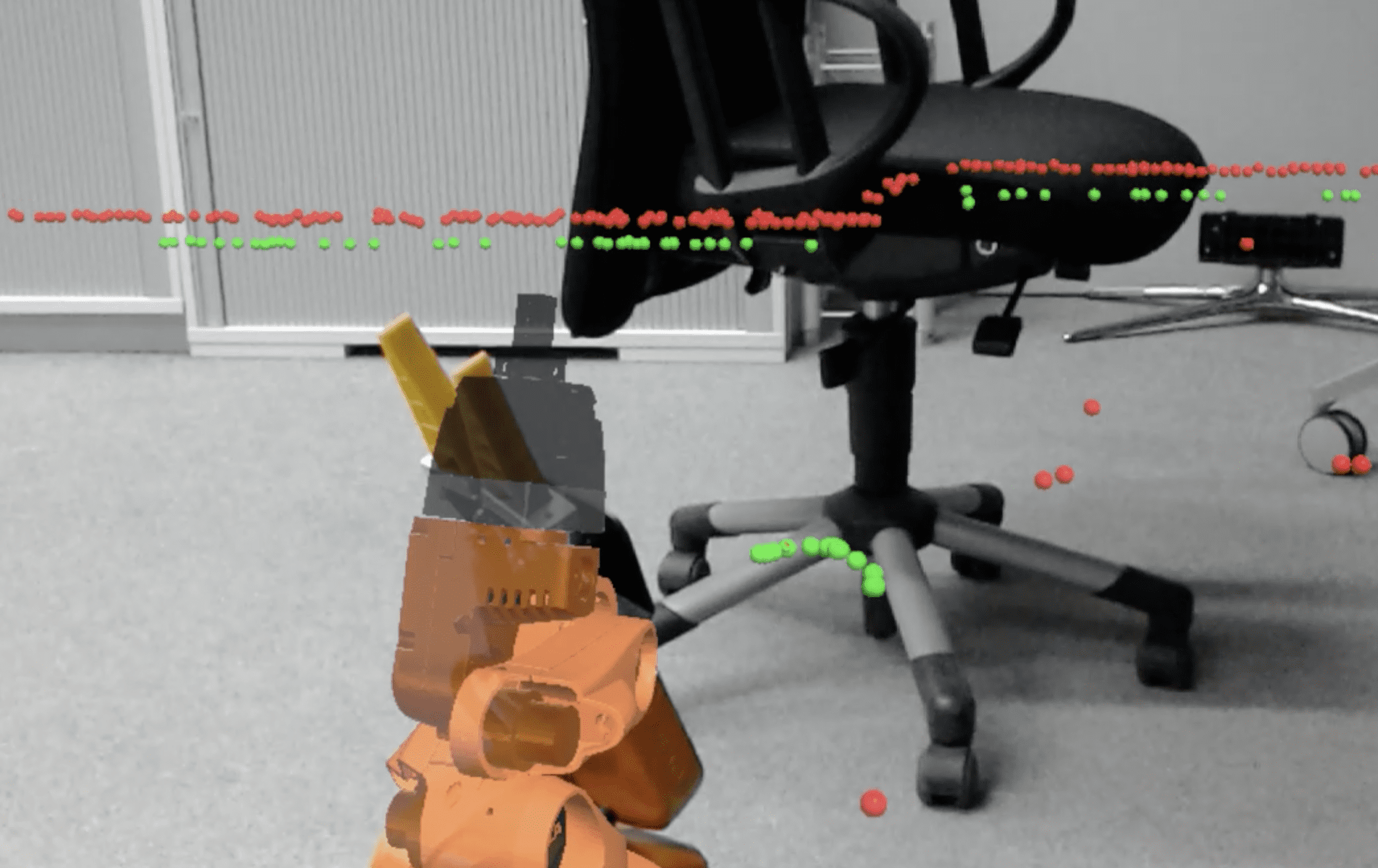}
	\caption{Laser scan (green) and environment map (magenta) visualization}
	\label{laser2}
\end{figure}

 Fig.\ref{laser2} is showing the visualization of laser scan data in green. Environment data is shown in red. The path information is visualized once the user defines a destination by dragging the blue arrow to a 3d position in the room. As fig. \ref{laser2} depicts, the laser scan data shapes nearby objects as well as far out walls, while the environment map (red) surrounding shapes of the whole room.
\section{EXPERIMENT}
To acquire relevant data for evaluation of performance aspects we defined 3 parameters to evaluate: first, frames per second (fps), second computational power, third, time to execution of robot movement. The above parameters were selected because they give relevant insight in the performance of the application while running the navigation visualization application. Frames per second is relevant because it shows the visual quality of the observed scene and if great amounts of data affect visual quality of the application. To evaluate computational power, we observe the CPU usage. Lastly, time to execution is an indicator of proper and fast data transmission between the two entities. This includes the time from defining the goal on the Hololens to the time when the path is visible to the user on the AR Headset. Our hypothesis is that if the application is demanding to much power, communication will also be slowed down.\\To acquire relevant data, we defined certain position with different views in order to get different robot poses to cover all possible laser scan setups. After that, the application is started and for every position, the above parameters are measured. Application CPU usage as well as fps can be accessed through developer Hololens portal. The time for execution of robot movement after the user defines a navigation goal is measured. For each measurement, 5 different visualization modes are used:
\begin{enumerate}
	\item Without any sensor data visualization
	\item With laser scan visualization
	\item With environment map visualization
	\item  With laser scan and environment map visualization
	\item With laser scan, environment and navigation visualization
\end{enumerate}
 The robot is driven to different position and each time the aforementioned parameters are collected and an average value is calculated. This is done a total of 20 times to get meaningful results. 
\section{EVALUATION}
In this section we evaluate the performance aspects of the Hololens for data visualization and communication between ROS and Hololens. We evaluate the performance of Hololens by displaying the different kinds of sensor data each for its own and afterwards simultaneously. Visual quality of the application is evaluated by observing fps.
For the overall performance of the Hololens, we evaluate the CPU usage for each visualization mode. To evaluate to what extinct the communication between the two entities is affected, the time to movement execution of the robot after defining a navigation goal on the Hololens is evaluated.
\subsubsection*{\textbf{2) (Frames per second)}}
As an important parameter for the visual performance of the Hololens, the fps was evaluated for different visualization modes. Results show, that fps is pending from 57 to 60 even when displaying laser scan and environment map data. 
This gives the conclusion that the Hololens is capable of visualizing huge amounts of scene objects without any visual downscale.

\subsubsection*{\textbf{2) (CPU usage)}}
\begin{figure}[h]
	\centering
	\includegraphics[width=3.3in]{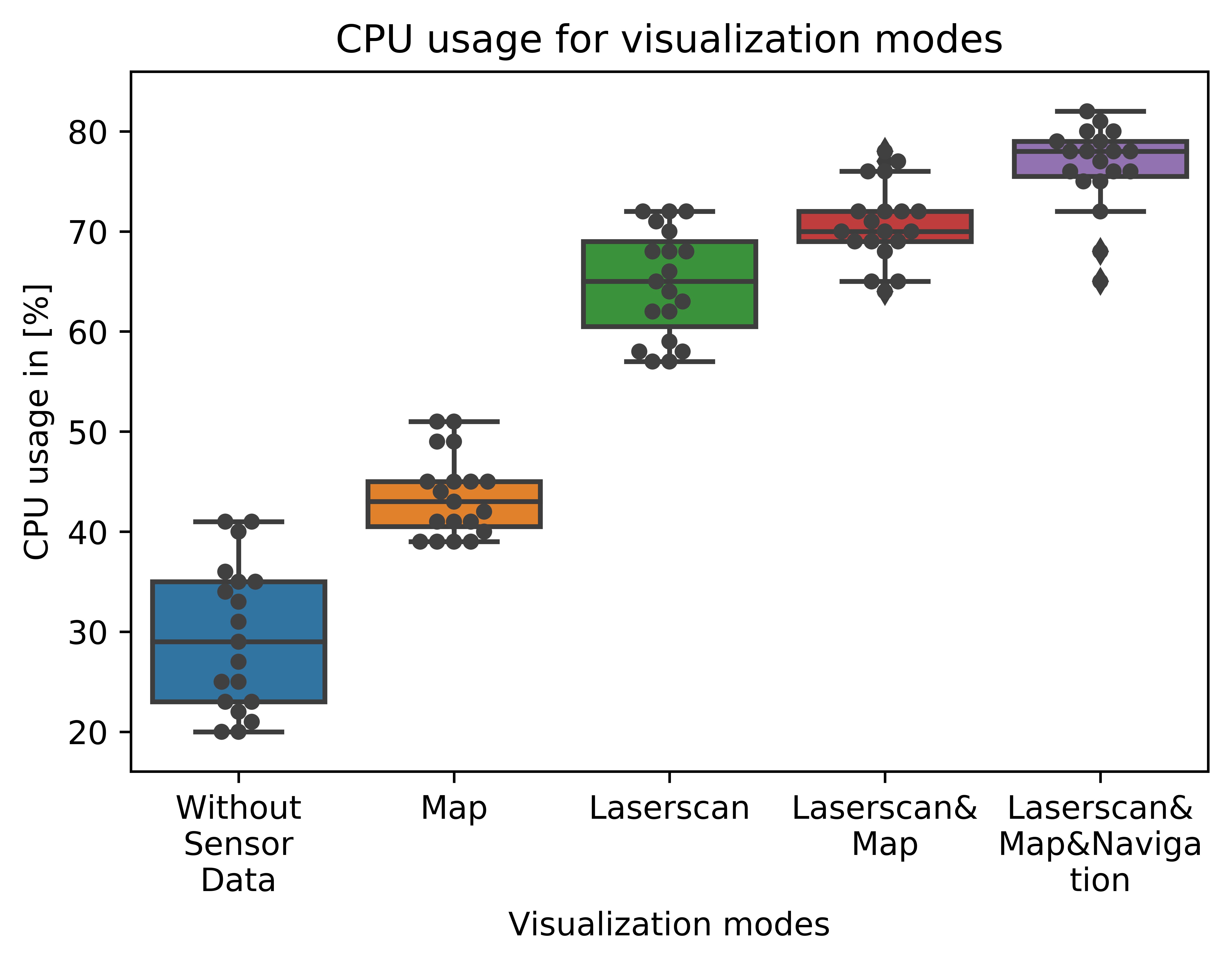}
	\caption{RAM usage for different visualization modes}
	\label{ram}
\end{figure}

The results in fig. \ref{ram} show a rise in CPU usage with more data being visualized. Without any visualization on, the CPU usage is pending on 20 to 30 percent. It is seen that CPU usage rises to 40 percent when map data is being visualized.  Display of laser scan data results in an increase to an average of 66 percent. CPU usage is highest when displaying Map, laser scan data and defining a navigation goal with an average of 79 percent. This is due to the spatial mapping of the whole environment which is necessary to define an exact 3D position of the room. The data transfer from Hololens to ROS is also demanding computational power. This results in a lag time between navigation goal placement and movement triggering for the robot. To evaluate this time lag the box chart in fig. \ref{move} illustrates the rising time to execution for the different visualization modes. 

\subsubsection*{\textbf{3) Time to Movement execution}}
\begin{figure}[h]
	\centering
	\includegraphics[width=3.3in]{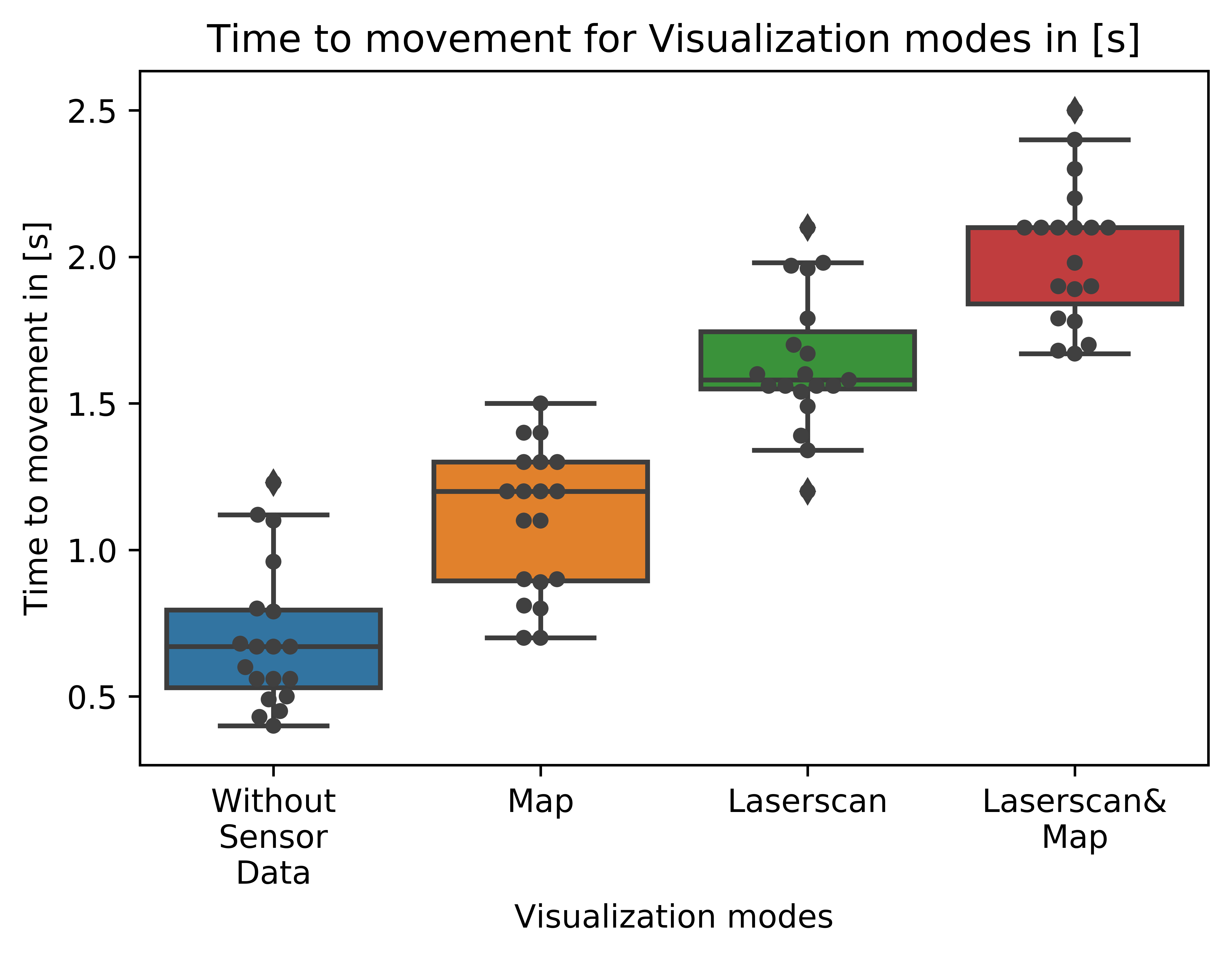}
	\caption{Time to movement execution for different visualization modes}
	\label{move}
\end{figure}

The box chart in fig. \ref{move} illustrates the duration between the user interaction and the robot movement for the different visualization modes. This includes the visualization of the navigation path planing data. Once again laser scan data has great affection on overall performance since the time increases dramatically sometimes also resulting in the crash of application. When no sensor data is visualized, average time for command execution is 0.5s. With mapping data added to display, time drops to 1.5s on average. Laser scan data displayed resulting in a rise to 1.8s and with laser scan and mapping data simultaneously on, to 2.2s. The last mode increases execution time to 2.6s.

\section{Discussion}
The fps throughout all visualization modes was constant and robust showing that the Hololens is capable of visualizing high amounts of game objects. Visualization of environment map do not considerably affect the fps and the CPU usage of the Hololens performance despite containing huge amounts of scene objects to be visualized. Also path information contain few scene objects to be visualized and do not have a great affection on frames per seconds and CPU usage.
However, the evaluation have shown that visualization of constantly incoming data clearly affects the overall Hololens performance. This is being proven by evaluation of CPU usage. There, only little increase in CPU usage have been observed when displaying the environment map although containing more scene objects than laser scan data. This is due to the fact that information of the map is only transferred and visualized once whereas laser scan data continuously change. The box plots clearly illustrate the rise in computational demand when displaying laser scan data. In the case of goal placement the overall performance drops even further due to the now additional required computational demands for communicating different information simultaneously. Furthermore, for goal definition spatial mapping is executed to extract the 3D location of the room. An considerably increased time have been observed when using goal definition together with data visualization. 

\section{CONCLUSION}
We demonstrated, that the Hololens is capable of visualizing important navigation stack sensor data within a cutting edge AR device. We propose a prototype application which successfully displayed key parameters of navigation stack within the Microsoft Hololens. The latter was evaluated in terms of performance to give relevant insight about the feasibility for integration of an HMD-AR into mobile robotics. Evaluation results show that the Hololens is capable of displaying a great amount of sensor data without any visual downscale. However, it is struggling for real-time data visualization. This is especially the case for visualization of laser scan data since the data flow is continuously changing and producing high amounts of data in parallel. This affects the overall performance of the application which is proved by the fact that goal definition takes considerably longer with laser scan visualization mode on. CPU usage was peaking at 80 percent when all sensor data were displayed simultaneously and navigation goal definition took considerably longer with sensor data on but did not affect the application crucially in terms of accuracy and robustness. Further work must include the optimization of preprocessing of the laserscan to reduce CPU usage.

%%%%%%%%%%%%%%%%%%%%%%%%%%%%%%%%%%%%%%%%%%%%%%%%%%%%%%%%%%%%%%%%%%%%%%%%%%%%%%%%

%%%%%%%%%%%%%%%%%%%%%%%%%%%%%%%%%%%%%%%%%%%%%%%%%%%%%%%%%%%%%%%%%%%%%%%%%%%%%%%%

%%%%%%%%%%%%%%%%%%%%%%%%%%%%%%%%%%%%%%%%%%%%%%%%%%%%%%%%%%%%%%%%%%%%%%%%%%%%%%%%

%%%%%%%%%%%%%%%%%%%%%%%%%%%%%%%%%%%%%%%%%%%%%%%%%%%%%%%%%%%%%%%%%%%%%%%%%%%%%%%%

\bibliographystyle{ieeetr}

\bibliography{references}

\end{document}